\documentclass[10pt,confrence]{IEEEtran}
\IEEEoverridecommandlockouts
\usepackage{cite}
\usepackage{amsmath,amssymb,amsfonts}
\usepackage{algorithmic}
\usepackage{graphicx}
\usepackage{textcomp}
\usepackage{xcolor}
\usepackage{subcaption} 
\usepackage{booktabs}
\usepackage[normalem]{ulem}
\usepackage{bbding}
\usepackage{url}
\usepackage{svg}
\usepackage{booktabs}
\usepackage{multirow}
\usepackage{float}
\usepackage{hyperref}
\usepackage[numbers]{natbib}
\def\BibTeX{{\rm B\kern-.05em{\sc i\kern-.025em b}\kern-.08em
    T\kern-.1667em\lower.7ex\hbox{E}\kern-.125emX}}

\usepackage{orcidlink}

\begin{document}

\title{CCS: Continuous Learning for Customized \\ Incremental Wireless Sensing Services

}

\author{Qunhang Fu\orcidlink{0009-0007-9789-0300},
        Fei Wang\orcidlink{0000-0002-0750-6990},~\IEEEmembership{Member,~IEEE,}
        Mengdie Zhu\orcidlink{0009-0007-3965-6700},\\
        Han Ding\orcidlink{0000-0002-5274-7988},~\IEEEmembership{Member,~IEEE,}
        Jinsong Han\orcidlink{0000-0001-5064-1955},~\IEEEmembership{Senior Member,~IEEE}
        Tony Xiao Han,~\IEEEmembership{Senior Member,~IEEE}
        \thanks{
        Qunhang Fu(email: 3122358189@stu.xjtu.edu.cn), Fei Wang, and Mengdie Zhu are with the School of Software Engineering, Xi'an Jiaotong University, Xi'an China. Han Ding is with the School of Computer Science and Technology, Xi'an Jiaotong University, Xi'an China. Jinsong Han is with the College of Computer Science and Technology, Zhejiang University, Hangzhou, China.  Tony Xiao Han is with the Wireless Technology Laboratory, Huawei Technologies Co. Ltd. Shenzhen, China. Fei Wang is the corresponding author (email: feynmanw@xjtu.edu.cn).
        }
        }

\markboth{IEEE Internet of Things Journal,~Vol.~00, No.~0, December~2024}%
{Shell \MakeLowercase{\textit{et al.}}: Bare Demo of IEEEtran.cls for IEEE Journals}

\maketitle

\begin{abstract}
Wireless sensing has made significant progress in tasks ranging from action recognition, vital sign estimation, pose estimation, etc. After over a decade of work, wireless sensing currently stands at the tipping point transitioning from proof-of-concept systems to the large-scale deployment. We envision a future service scenario where wireless sensing service providers distribute sensing models to users. During usage, users might request new sensing capabilities. For example, if someone is away from home on a business trip or vacation for an extended period, they may want a new sensing capability that can detect falls in elderly parents or grandparents and promptly alert them. In this paper, we propose CCS (continuous customized service), enabling model updates on users' local computing resources without data transmission to the service providers. To address the issue of catastrophic forgetting in model updates—where updating model parameters to implement new capabilities leads to the loss of existing capabilities—we design knowledge distillation and weight alignment modules. These modules enable the sensing model to acquire new capabilities while retaining the existing ones. We conducted extensive experiments on the large-scale XRF55 dataset across Wi-Fi, millimeter-wave radar, and RFID modalities to simulate scenarios where four users sequentially introduced new customized demands. The results affirm that CCS excels in continuous model services across all the above wireless modalities, significantly outperforming existing approaches like OneFi~\cite{onefi}. 

\end{abstract}


\section{Introduction}\label{sec:introduction}
After more than a decade of research, wireless sensing is now at a tipping point, transitioning from laboratory settings to large-scale deployment. Several developments have attested to this shift, including the consideration of integrated sensing and communication in the IEEE 802.11bf protocol standardization~\cite{802}, Google's development of Soli millimeter-wave radar chips for gesture recognition in the Pixel smartphones~\cite{soli}, and the emergence of numerous wireless sensing startups like ORIGIN Wireless. Perceiving the tipping point, we proactively delve into a prospective business schema for wireless sensing, in which two main participants are involved: wireless sensing service providers and users. Wireless sensing service providers initially distribute trained wireless sensing models to users, constituting their base service. Subsequently, during daily lives, users may encounter ongoing personalized requirements that the initial sensing models do not cater to. For instance, when a user needs to be away for an extended period, s/he may desire new sensing capability to detect falls in elderly parents or grandparents. If service providers can also respond to such customized needs, it will constitute another significant revenue stream. 

\begin{figure}[t]
    \centering
    \includegraphics[width=1\linewidth]{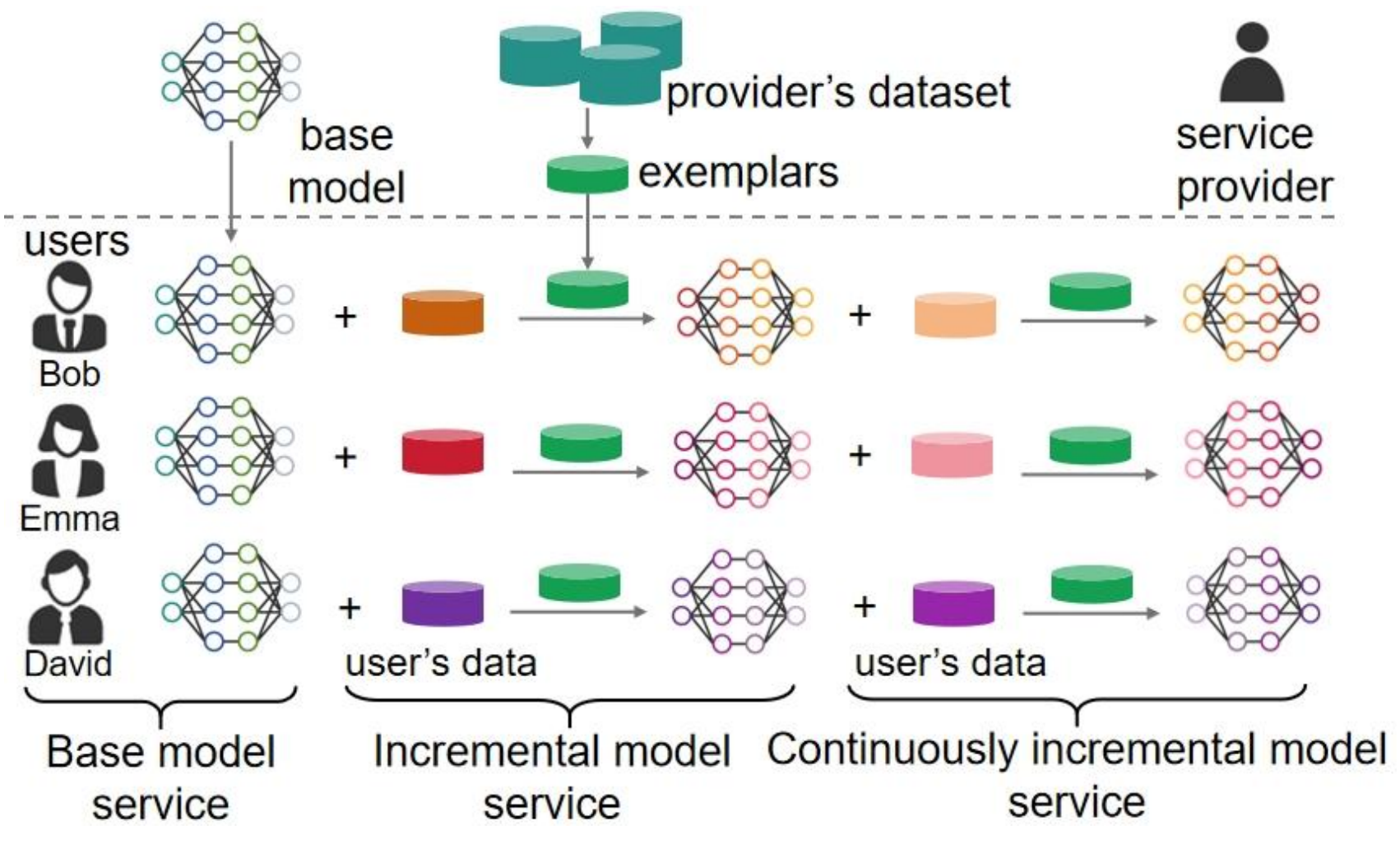}
    \caption{We present CCS. With CCS, the service provider can not only provide base wireless sensing service for users but also provide incremental model service to meet users' continuous customized sensing demand under the premise that users' wireless sensing data is never transmitted to the service provider to ensure data privacy protection.}
    \label{fig:ccs}
    
\end{figure}

The simplest solution would be for users to send their data to the service providers, whereupon providers retrain a model tailored to the new requirements and then return it to users. However, this approach risks compromising user data privacy. One privacy-preserving solution proposed is federated learning, a technique that allows users to transmit only parameters associated with the raw data to service providers, rather than the data itself~\cite{fl1,fl2,fl3}.  However, some literature has pointed out that such parameters can be leveraged to reconstruct the original data~\cite{fl4,fl5}.  Moreover, the above approaches may also raise network strain and computational burdens on the service providers, particularly as the user base expands significantly.


This paper introduces CCS, a scheme enabling service providers to attain the capacity for delivering such continuous services. The design of CCS is predicated on a fundamental assumption: to safeguard data privacy, user wireless sensing data is never transmitted to the service provider. As depicted in Fig.~\ref{fig:ccs}, the implementation of this scheme is structured into three distinct stages: base model service, incremental model service, and continuously incremental model service. In the first stage, the service provider distributes a wireless sensing model to users, enabling them to access wireless sensing services locally on their devices, even in situations where network connectivity is lost. 
As user requirements evolve, CCS facilitates the introduction of new sensing demands through the provision of new data by the users. At this stage, CCS requires users to acquire wireless data that supports the new requirements and store it locally. Service providers then leverage this data, harnessing the computational resources of the user's device, to update the model, thereby endowing it with the capability to meet the new demands. CCS supports not only singular model updates but also the ongoing evolution of model services, which we denominate as the continuously incremental model service, shown in Fig.~\ref{fig:ccs}.

The issue of catastrophic forgetting~\cite{cf1,cf2,cf3} often arises during model updates in incremental model services. This occurs when the model is updated with users' new data to meet new requirements, causing it to lose its existing capabilities. To maintain recognition accuracy for existing data categories while incorporating new ones, the service provider could select a subset of data from its repository, called exemplars\cite{ex1,ex2}, which would then be transmitted to the users' local devices for participation in model updating, as illustrated in Fig.\ref{fig:ccs}. Given that data is the vital asset for service providers, exemplars would be kept as compact as feasible. CCS leverages Herding techniques~\cite{Herding} to select exemplars, from the provider's data repository, and deliver the exemplars to users. 


Furthermore, CCS initializes the current stage model $\mathcal{M}_{s}$ with model from the previous stage, $\mathcal{M}_{s} \gets \mathcal{M}_{s-1}$. $\mathcal{M}_{s}$ is updated with new user's data and exemplars from the provider. Meanwhile, the parameters of $\mathcal{M}_{s-1}$ are frozen as teacher network and distilled into $\mathcal{M}_{s}$ as knowledge of the old service, ensuring that $\mathcal{M}_{s}$ does not forget the previous service during updates. Moreover, CCS applies weight aligning techniques~\cite{wa} to balance the model's propensity towards the new and old services. As users continue to acquire new services,  $\mathcal{M}_{s}$ will serve as the initialization for the model in the next stage, $\mathcal{M}_{s+1} \gets \mathcal{M}_{s}$, with these operations iteratively executed to achieve continuously incremental model service.

We use the large-scale XRF55 dataset~\cite{xrf55} to evaluate CCS. XRF55 encompasses 55 indoor action classes spanning human-computer interaction, human-human interaction, and human-object interaction, incorporating three wireless modalities: Wi-Fi, millimeter-wave radar, and RFID. During evaluation, we initially partition XRF55 into five subsets, each consisting of 15, 10, 10, 10, and 10 action categories, respectively. We posit that the service provider trains and distributes a base model using the subset containing 15 actions. Subsequently, the user progressively adds four incremental action recognition requirements of 10 classes each, employing the CCS methodology. We introduce new demand categories in various sequences across Wi-Fi, millimeter-wave radar, and RFID. Extensive experiments reveal that CCS effectively ensures model competence in recognizing newly added action categories while largely preserving its ability to identify previously learned action classes. To assess model worth, we adopt a novel metric – ACCN, the multiplication of the number of recognizable action categories and model accuracy, which can balance the model capacity and accuracy. Our validation demonstrates that this metric consistently improves with each incremental model service stage and significantly outperforms existing approaches, such as iCaRL~\cite{icarl}, UCIR~\cite{ucir}, BiC~\cite{bic}, OneFi~\cite{onefi}. We summarize the contributions of this paper as follows.

(1) We have contemplated a potential future business schema for wireless sensing services, encompassing service providers and users. For the first time, we have addressed the issue of how service providers might accommodate users' continuously emerging new requirements, proposing the CCS solution in this regard.

(2) The proposed CCS encompasses three key technical aspects: first, the selection of exemplars; second, model updating via knowledge distillation; and third, the alignment of weights for new and old services. Collectively, these techniques ensure that while CCS provides new services, it does not discard the existing services.

(3) We have conducted extensive comparative studies and ablation experiments on the WiFi, millimeter-wave radar, and RFID modalities within the XRF55 dataset. The results demonstrate that the proposed CCS effectively mitigates catastrophic forgetting issues encountered during continuous model service.

\section{RELATED WORK}\label{sec:related_work}

\subsection{Few-shot Learning}

Few-shot learning allows for updating deep models with several new samples, which may provide incremental demands from users in wireless sensing model services. For example, OneFi\cite{onefi} stands as a representative work of few-shot learning, with its core concept being the ability to recognize unseen gestures using only one (or a few) labeled samples. This method, after learning 20 base classes in the pre-training stage, can rapidly learn new classes never seen before with a small amount of data samples through fine-tuning. OneFi primarily emphasizes how to quickly adapt to new classes, but it can not recognize old action classes. In contrast, our CCS effectively preserves and updates the knowledge of old action classes by introducing a memory mechanism during model training, thus it is suitable for continuously incremental model services.

Another work, Wi-Fringe\cite{wifringe}, is a device-free human activity recognition system based on WiFi data. Leveraging the principle of zero-shot learning, it combines radio frequency signal features with attribute and context-aware representations of English words or phrases, enabling inference of activities without prior training samples. While Wi-Fringe can identify some unseen activity categories, this capability relies on the similarity of label texts and cannot recognize new categories that lack semantic relevance with previous categories in text. This implies a limitation of Wi-Fringe when dealing with new categories significantly different from previous ones in terms of semantics, as it cannot utilize existing text information to infer activities of these new categories. In contrast, our proposed CCS does not rely on text information. Utilizing the method of class-incremental learning, it addresses catastrophic forgetting while accommodating the learning of new category demands. This enables CCS to better meet the needs of different users without being restricted by the similarity of text information, while maintaining good generalization capability.

\subsection{Continuous Learning}
Continuous learning or incremental learning aims for ``learning without forgetting''~\cite{lwf}, which can be utilized to provide incremental demands from users in wireless sensing model services. Regularization\cite{regular1,regular2,regular3,regular4} and data replay\cite{replay1,regular1,replay3} are two common incremental learning methods in human action recognition tasks.  One representative regularization method is knowledge distillation\cite{lwf,kd1,kd2}. Its core concept involves introducing an additional loss function to guide the training of the student model using the knowledge from the teacher model, facilitating better learning and generalization. The goal of this process is to enable the student model to mimic the behavior of the teacher model as closely as possible. It not only employs the loss function of the true labels but also introduces an extra loss function to measure the similarity between the predictions of the student and teacher models, often using cross-entropy loss or other forms of distance metrics. Since the regularization method does not store existing activity samples, its recognition accuracy for existing activities is weak. iCaRL~\cite{icarl} represents the data replay method and, in addition to employing knowledge distillation, also selects a subset of samples from known classes in each incremental stage and incorporates them into the exemplar set along with the data from newly added classes for training. For the selection of exemplar samples, random selection is the most primitive method, and Herding~\cite{Herding} has become a commonly used strategy for selecting exemplars. In this paper, we also adopt the Herding strategy for selecting exemplar samples.

\section{Methodology}\label{sec:methodology}
\subsection{Problem State}\label{sec:problem_state}

\begin{figure*}[t]
    \centering
    \includegraphics[width=0.8\textwidth]{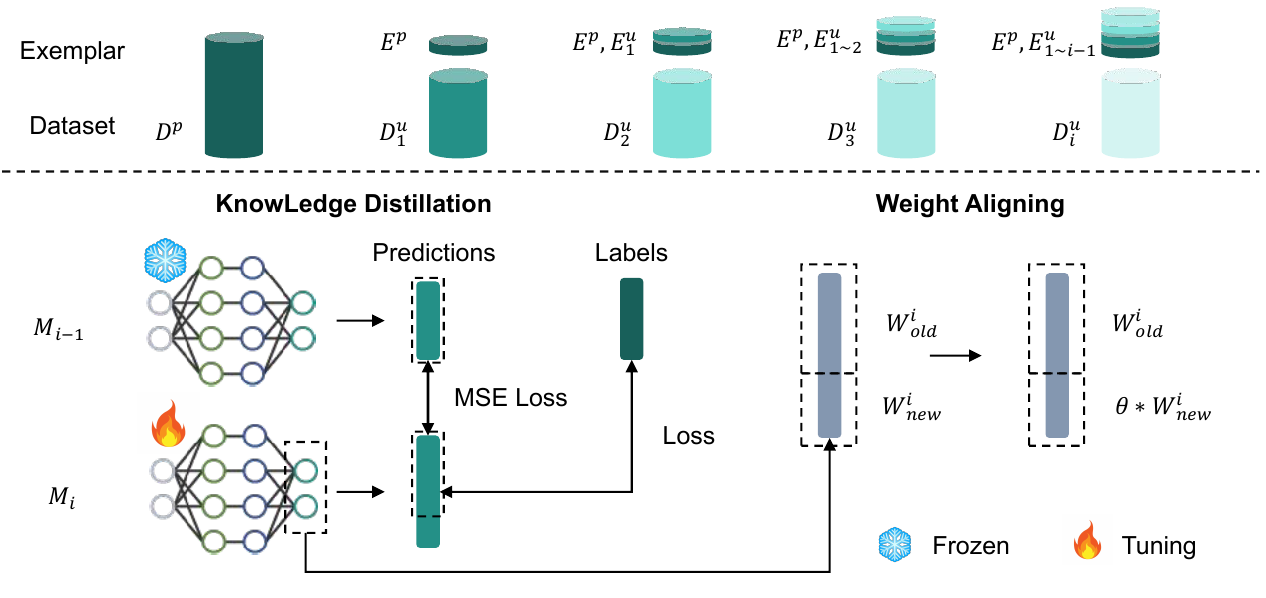}
    \caption{We incorporate knowledge distillation and weight aligning throughout the training process of each incremental stage. During each incremental stage, exemplars are extracted from the training data of the previous stage using the Herding method~\cite{Herding}. The network learns to retain knowledge of old tasks by employing knowledge distillation from a frozen model of the previous stage, utilizing MSE loss for distillation. Additionally, weight aligning addresses the issue of imbalance in the weight distribution between new and old services.}
    \label{fig:layout}
\end{figure*}

As shown in Fig.~\ref{fig:layout}, CCS enables wireless sensing  service providers provide base model service, incremental model service, and continuously incremental model service for users. We term these service stages as $S_i, i\in [0, n]$, where $S_0$ represents the base model service; $S_i$($i>0$) represents the incremental service at the $i$-th stage. In $S_0$, the service provider trains a base model $\mathcal{M}_0$ with the dataset from the provider $D^p$, and distribute to users' local devices as the base model service. 

As the first incremental model services, a user generate new demands. To align $\mathcal{M}_0$ with these needs, CCS necessitates that users locally prepare data, $D_1^u$, relevant to the new demand for updating $\mathcal{M}_0$ . If $\mathcal{M}_0$ were solely updated with $D_1^u$, it would likely suffer from catastrophic forgetting, losing its initial capabilities. Addressing this issue, CCS requires the service provider to select a subset of data (exemplars) $E^p$ from $D^p$, to be transmitted to the user's local device for inclusion in the model update process, yielding $\mathcal{M}_1$. The objective of CCS is to ensure that $\mathcal{M}_1$ not only accommodates the user's new requirements but also retains its original capabilities. This process can be represented as Equation.~\ref{eq:m1}. 
\begin{equation}\label{eq:m1}
    \mathcal{M}_1 = \text{CCS}(\mathcal{M}_0,  D_1^u, E^p)
\end{equation}

Following this logic, when updating $\mathcal{M}_i$($i\ge2$), the process can be represented as Equation.~\ref{eq:ms}. 
\begin{equation}\label{eq:ms}
    \mathcal{M}_i = \text{CCS}(\mathcal{M}_{i-1}, D_i^u, E^p, E_1^u, E_2^u,...,E_{i-1}^u), i\ge2
\end{equation}
where $D_i^u$ represents user's dataset for new demands in stage $S_{i}$; $E_{i-1}^u$ represents exemplars from $D_{i-1}^u$. 

Next, we will introduce how to select exemplars in Sec.~\ref{sec:exemplar} and how to update models in Sec.~\ref{sec:knowledge} and Sec.~\ref{sec:weight} in detail.

\subsection{Exemplar Selection}\label{sec:exemplar}

To mitigate catastrophic forgetting in continuous learning, selecting exemplars in previous learning stage has been widely demonstrated as an effective method~\cite{rainbow,gradient,anchor}. For example, one can choose the centroids of K-Means clustering as exemplars. In this paper, we apply the more advanced Herding method~\cite{Herding} to select exemplars. Suppose the provider's data $D^p$ includes $C$ classes of data for services, for $c$-th class, its data sample is denoted as $d_i^c, i\in[1,N_c]$, where $N_c$ represents the total number of data samples of the $c$-th class. We compute the normalized features of $d_i^c, i\in[1,N_c]$, and take the average value as the feature center for the $c$-th class of the provider's data. 
\begin{equation}\label{eq:featurecenter}
F_c^p = \frac{1}{N_c} \times \sum_{i=1}^{N_c} \text{Norm}(\mathcal{M}_0(d_i^c))
\end{equation}
We select the $K$ data samples that are nearest (Euclidean distance) to this feature center as exemplar subset $E_c^p$.
By repeating the above process for all classes within the provider's data $D^p$, we can acquire the exemplars $E^p$. Therefore, $E^p$ includes $C\times K$ samples. A smaller K indicates that the selected exemplars are more compact or sparse. The exemplars $E^p$ then participates in updating $\mathcal{M}_1$ in Equation.~\ref{eq:m1} for the incremental model service. 

Following this logic, we can select user's exemplars $E_{i-1}^u$ at stage $S_i$ for continuously incremental model service, shown in Equation.~\ref{eq:ms}. The upper part of Fig.~\ref{fig:layout} illustrates how the number of exemplars involved in model updates accumulates over each stage. 

\subsection{Knowledge Distillation}\label{sec:knowledge}

Knowledge distillation involves training a student model $\mathcal{M}_i$ to mimic and learn from a teacher model $\mathcal{M}_{i-1}$, which has already acquired knowledge of old knowledge. This process not only mitigates catastrophic forgetting during the learning of new knowledge but also effectively leverages previously acquired knowledge while maintaining overall model performance. This section details how the model $\mathcal{M}_i$ is updated in knowledge distillation, utilizing the user data corresponding to the new demands of the current stage ($D_i^u$), the model from the previous stage $\mathcal{M}_{i-1}$, and the set of exemplars. The ultimate objective of the method is to ensure that $\mathcal{M}_i$, while meeting the new demands, also ideally retains the capabilities of serving the older functions.

As illustrated in Fig.~\ref{fig:layout}, CCS initially sets $\mathcal{M}_{i-1}$ as the starting point for $\mathcal{M}_i$ and freezes $\mathcal{M}_{i-1}$ to act as the teacher network for $\mathcal{M}_i$. It is noteworthy that, during the $i$-th stage, new services are added; consequently, prediction head of $\mathcal{M}_{i-1}$ expands to accommodate services for the new task. The loss function to updates the model is as follows.
\begin{equation}\label{eq:updateloss}
Loss = (1-\alpha)  \text{Dis}(\mathcal{M}_i(x), y) + \alpha  \text{MSE}( \mathcal{M}_i(x), \mathcal{M}_{i-1}(x))
\end{equation}
where $(x,y)$ represents a pair consisting of a training sample and its corresponding label. $\text{Dis}(\mathcal{M}_i(x), y)$  represents the distance between the prediction of $\mathcal{M}_s$ and the true value, responsible for enabling the model to acquire new service capabilities. $\text{MSE}( \mathcal{M}_i(x), \mathcal{M}_{i-1}(x))$  represents the MSE loss between output of $\mathcal{M}_i$ and $\mathcal{M}_{i-1}$, distilling knowledge from $\mathcal{M}_{i-1}$ to $\mathcal{M}_i$. $\alpha$ is to balance the two distances.

\subsection{Weight Aligning} \label{sec:weight}

Introducing new service tasks can disrupt model eight allocation, causing model imbalance and reducing learning capacity. This can lead to suboptimal performance on new tasks and catastrophic forgetting of previously learned tasks. In response to these challenges, we introduces additional weight adjustment mechanisms during model training to effectively balance the distribution of weights between new and old tasks, described next.

Assuming $M_{i-1}$ is capable of providing $u$ old services, and $M_{i}$ introduces $v$ new services, the weights of  prediction head of $M_{i}$ can accordingly be represented as Equation.~\ref{eq:weight1}.
\begin{equation}\label{eq:weight1}
W^{i} = (w_1,w_2,...,w_{u},w_{u+1},...,w_{u+v}  ). 
\end{equation}

We leverage weight aligning technique~\cite{wa} to re-balance the weights between old services and  new services as Equation.~\ref{eq:weight_align}.

\begin{equation}\label{eq:weight_align}
\begin{gathered}
\text{Norm}_{old} = (\left\|w_1\right\|,\left\|w_2\right\|,...,\left\|{w_u}\right\|)\\
\text{Norm}_{new} = (\left\|w_{u+1}\right\|,...,\left\|w_{u+v}\right\|)\\
(\widehat{w}_{u+1},...,\widehat{w}_{u+v}) = \frac{ \text{Mean} (\text{Norm}_{old} )}{ \text{Mean}(\text{Norm}_{new} )} \times (w_{u+1},...,w_{u+v}  )
\end{gathered}
\end{equation} 
where $\widehat{w}_{u+i},i\in[1,v]$ represent the weights for new services after weight aligning. After weight aligning, the weights of prediction head of model $M_{i}$ turns to be
\begin{equation}\label{eq:weight_model-i}
\widehat{W}^{i} = (w_1,w_2,...,w_{u}, \widehat{w}_{u+1},...,\widehat{w}_{u+v} ). 
\end{equation}

\subsection{Hyper-parameter Setting} 

For each new service class added, we select one exemplar, thus setting K = 1, mentioned in  Sec.~\ref{sec:exemplar}. We set $ \alpha = 0.1\times u/(u+v)$ in Equation.~\ref{eq:updateloss} to balance the two distances for model updating. We set $\text{Dis}()$ as cross-entropy loss for the classification task.

\section{Experiments}\label{sec:experiments}

\subsection{Evaluation Dataset}\label{sec:dataset}

We use the large-scale XRF55 dataset~\cite{xrf55} to evaluate CCS, primarily for two reasons: (1) The XRF55 dataset encompasses three distinct wireless modalities – Wi-Fi, millimeter-wave radar, and RFID – allowing us to test CCS's adaptability and versatility across different wireless modalities; (2) Comprising 55 categories of action data, XRF55 boasts the broadest range of action classes currently available, facilitating the continuous addition of new classification tasks as users' needs, which in turn enables a comprehensive assessment of CCS's capability for incremental model services and continuously incremental model service. We experimentally employ ResNets~\cite{he2016deep} as the sevice models for RFID and Wi-Fi modalites, and empirically employ Temporal UNet~\cite{wang2019temporal} as the service model for mmWave radar modality.

\begin{table}[t]
\centering
 \setlength{\tabcolsep}{2pt} 
\resizebox{\columnwidth}{!}{%

\begin{tabular}{ccccc}
\hline
Action & Person & Modality & Training samples & Test samples \\ \hline
55 & 30 & Wi-Fi, mmWave, RFID & 23100 & 9900  \\ \hline
\end{tabular}
}
\caption{The XRF55 dataset~\cite{xrf55} encompasses 55 distinct actions across three wireless modalities. We use its samples at the scene \#1, totaling 23100 training samples and 9900 testing samples.}
\label{tab:xrf55}
\end{table}

\begin{table}[t]
\centering
 \setlength{\tabcolsep}{2.5pt} 
\resizebox{\columnwidth}{!}{%
\begin{tabular}{cccccccc}
\toprule
      & Stage 1 & Stage 2 & Stage 3 & Stage 4 & Stage 5 & Result table   & Result figure \\
      \midrule
User1 & C1$\sim$C15  & C16$\sim$C25 & C26$\sim$C35 & C36$\sim$C45  & C46$\sim$C55  & Table.~\ref{tab:result1} & Fig.~\ref{fig:result1} \\
User2 & C1$\sim$C15  & C26$\sim$C35  & C46$\sim$C55  & C36$\sim$C45  & C16$\sim$C25  & Table.~\ref{tab:result2}      & Fig.~\ref{fig:result2} \\
User3 & C1$\sim$C15  & C36$\sim$C45  & C46$\sim$C55  & C16$\sim$C25  & C26$\sim$C35  & Table.~\ref{tab:result3}      & Fig.~\ref{fig:result3} \\
User4 & C1$\sim$C15  & C46$\sim$C55  & C16$\sim$C25  & C26$\sim$C35  & C36$\sim$C45  & Table.~\ref{tab:result4}      & Fig.~\ref{fig:result4}\\

\bottomrule
\end{tabular}
}
\caption{ We assume four users require customized incremental services in differing sequences. We assess CCS under this configuration. The results are shown in corresponding tables and figures as listed.}
\label{tab:user_sequences}
\end{table}

To simulate the base model service in CCS and the scenario where users continuously add new requirements, we randomly divided the XRF55 dataset into five subsets based on action categories, with each subset containing 15, 10, 10, 10, and 10 classes of action data, respectively. The subset with 15 classes is utilized to train the base model in CCS, while the remaining subsets are sequentially introduced as new action recognition demands, serving to assess the capability of CCS's incremental model services and continuously incremental model service. For simplicity, we denote the first subset, which includes action categories 1 through 15, as C1$\sim$C15. The second subset, comprising categories 16 to 25, is denoted C16$\sim$C25. Following this pattern, the remaining three subsets, encompassing subsequent action categories, are correspondingly named as C26$\sim$C35, C36$\sim$C45, and C46$\sim$C55, respectively.

Table.~\ref{tab:user_sequences} illustrates our evaluation scheme, in which we assume four users requiring customized incremental services in differing sequences. Samples of C1$\sim$C15 serve as the base model service in the first stage, and the remaining subsets representing incremental services. Under this configuration, we assess the performance of CCS, and report the results in upcoming figures and tables of this section, specified in Table.~\ref{tab:user_sequences}. 
In stages 2-5, we employ the Herding~\cite{Herding} method to select exemplars from the preceding stage, at a ratio of one sample per person per action, for model updating. For instance, in stage 2, we have $1\times30\times 15=450$ exemplars to update models, which is $1/14\approx 7\%$ of training data in stage 1.



\subsection{Evaluation Metrics}

\textbf{(1) Accuracy } We use classification accuracy to indicate CCS's ability to classify actions at different service stages. 


\begin{table}[t]
     \setlength{\tabcolsep}{2pt} 
    \centering
    \resizebox{\columnwidth}{!}{%
    \begin{tabular}{llccccc}
        \toprule
        \textbf{User1} & \textbf{Method} & \textbf{C1$\sim$C15} & \textbf{+C16$\sim$C25} & \textbf{+C26$\sim$C35} & \textbf{+C36$\sim$C45} & \textbf{+C46$\sim$C55} \\
        \midrule
        \multirow{5}{*}{RFID} & Baseline & 75.96 & 32.82 & 20.24 & 17.42 & 13.43 \\
             & iCaRL & 75.85 & 48.87 & 33.29 & 30.54 & 26.67 \\
             & UCIR & 74.30 & 49.38 & 34.51 & 31.94 & 27.71 \\
             & BiC & 73.81 & 55.02 & 36.56 & 24.31 & 23.55 \\
             & OneFi & 58.81 & 14.44 & 7.21 & 5.75 & 6.47 \\
             & Ours & 75.96 & 58.56 & 44.89 & 38.23 & 34.01 \\
        \midrule
        \multirow{5}{*}{Wi-Fi} & Baseline & 95.07 & 39.09 & 26.54 & 20.86 & 18.33 \\
              & iCaRL & 94.85 & 72.51 & 65.86 & 56.01 & 56.57 \\
              & UCIR & 94.89 & 79.4 & 71.37 & 58.85 & 60.2 \\
              & BiC & 93.37 & 74.13 & 58.63 & 52.58 & 53.67 \\
              & OneFi & 84.30 & 20.24 & 10.40 & 10.32 & 8.64 \\
              & Ours & 95.07 & 83.56 & 75.48 & 64.12 & 65.10 \\
        \midrule
        \multirow{5}{*}{mmWave} & Baseline & 91.11 & 38.29 & 25.24 & 20.47 & 17.47 \\
               & iCaRL & 91.70 & 77.58 & 68.33 & 58.48 & 56.58 \\
               & UCIR & 91.41 & 80.60 & 72.38 & 64.48 & 63.45 \\
               & BiC & 91.30 & 79.29 & 73.33 & 66.59 & 59.99 \\
               & OneFi & 84.37 & 31.87 & 15.13 & 12.27 & 11.97 \\
               & Ours & 93.07 & 87.49 & 81.16 & 69.35 & 69.19 \\
        \bottomrule
    \end{tabular}
    }
    \vspace{2pt}
    \caption{
    Accuracy of CCS services for user1 with the continuous demands listed in Table.~\ref{tab:user_sequences}. CCS outperforms existing alternative methods such as iCaRL~\cite{icarl}, UCIR~\cite{ucir}, BiC~\cite{bic} and OneFi~\cite{onefi} by a large margin for three wireless modalities. 
    }
    \label{tab:result1}
    
\end{table}

\begin{figure}[t]
  \centering
  \includegraphics[width=\linewidth]{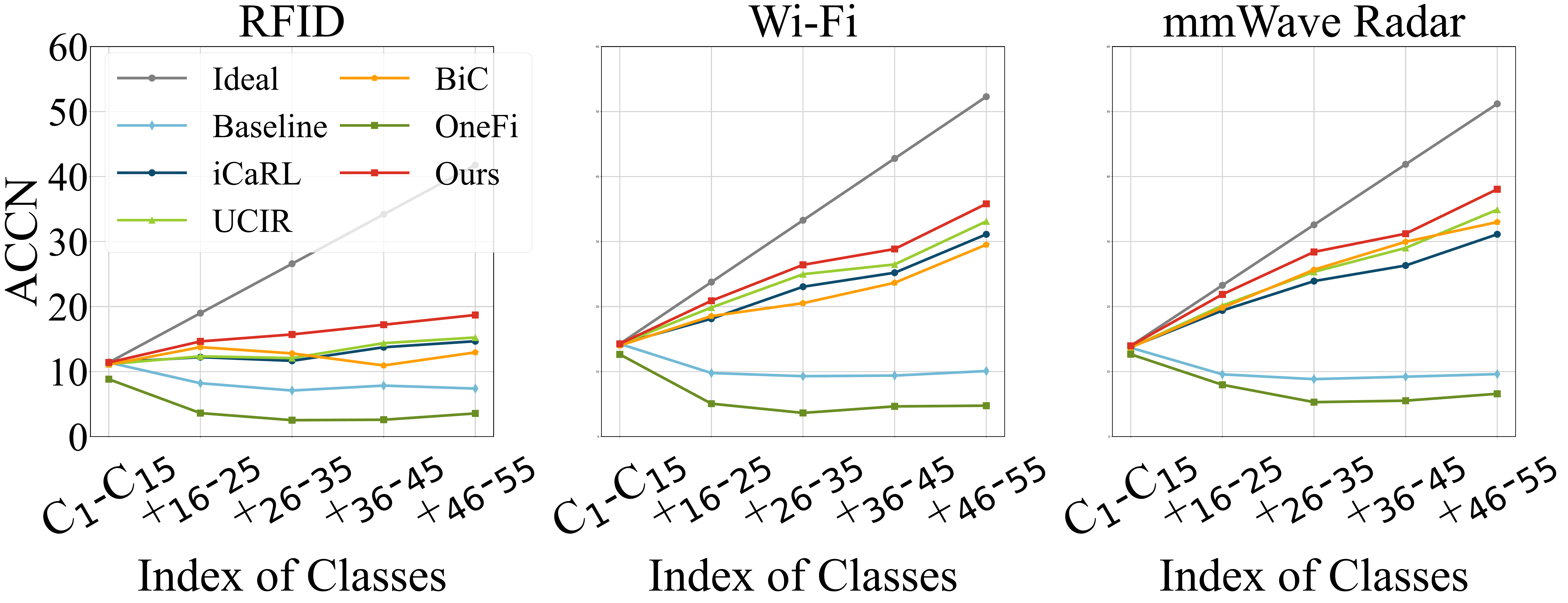}
  \caption{ 
  ACCN of CCS services for user1 with continuous stages listed in Table.~\ref{tab:user_sequences}. CCS traces the closest path to the ideal curve compared to alternative methods such as iCaRL~\cite{icarl}, UCIR~\cite{ucir}, BiC~\cite{bic} and OneFi~\cite{onefi}, demonstrating a consistent rise in model value.
  } 
  \label{fig:result1}
  
\end{figure}
\textbf{(2) ACCN } CCS's mode capacity, the number of recognizable action categories, would increase during incremental service, however, the average accuracy might decline during the forgetting old actions in incremental learning. We adopt a novel metric to assess model worth – the multiplication of the number of recognizable action categories and model accuracy~(ACCN), which can balance the model capacity and accuracy. 
\begin{equation}\label{eq:accn}
ACCN = N \times Accuracy
\end{equation}
where $N$ is the number of recognizable action categories by CCS models.


\subsection{Results}\label{sec:results}

\textbf{(1) Accuracy and ACCN on User1}

In Table.~\ref{tab:result1}, we show CCS's accuracy on user1 at five continuous stages listed in Table.~\ref{tab:user_sequences}. The baseline accuracy is derived from updating models with only new data in the new stage, which cannot recognize old actions even starting at the second stage.  It is observable that CCS exhibits high accuracy in the first stage, particularly with Wi-Fi and RFID data. As the number of stages increases, the accuracy gradually declines. By the final stage, the respective accuracy for RFID, Wi-Fi, and millimeter-wave radar are 34.01\%, 65.10\%, and 69.19\%, respectively. Nonetheless, it remains notably higher than current alternative methods like iCaRL~\cite{icarl}, UCIR~\cite{ucir}, and BiC~\cite{bic} by a large margin.  CCS also largely outperforms OneFi~\cite{onefi}, which is proposed to recognize unseen actions with one/few wireless samples. OneFi focus on enabling the model to quickly transfer and learn new action knowledge, without addressing the issue of catastrophic forgetting. As a result, OneFi almost completely loses its ability to recognize old actions during the incremental learning stage.





Fig.~\ref{fig:result1} shows ACCN curves, where the ideal scenario would be one where the model learns new information without forgetting old information; hence, the ACCN curve would linearly increase with the incremental capacity of the model. In reality, however, models partially forget previously learned knowledge. As depicted, all methods fall short of this ideal, yet, CCS, traces the closest path to this ideal curve, demonstrating a consistent rise in model value measured by ACCN.

\begin{table}[t]
     \setlength{\tabcolsep}{2pt} 
    \centering
    \resizebox{\columnwidth}{!}{%
    \begin{tabular}{llccccc}
        \toprule
        \textbf{User2} & \textbf{Method} & \textbf{C1$\sim$C15} & \textbf{+C26$\sim$C35} & \textbf{+C46$\sim$C55} & \textbf{+C36$\sim$C45} & \textbf{+C16$\sim$C25} \\
        \midrule
        \multirow{5}{*}{RFID} & Baseline & 72.96 & 27.40 & 20.83 & 16.81 & 14.58 \\
             & iCaRL & 74.15 & 43.84 & 29.89 & 26.98 & 21.43 \\
             & UCIR & 73.70 & 42.24 & 30.84 & 29.93 & 25.03 \\
             & BiC & 73.81 & 37.96 & 31.73 & 25.07 & 23.84 \\
             & OneFi & 58.81 & 10.87 & 10.05 & 5.74 & 5.74 \\
             & Ours & 74.30 & 52.31 & 39.83 & 34.64 & 32.14 \\
        \midrule
        \multirow{5}{*}{Wi-Fi} & Baseline & 95.07 & 37.02 & 28.16 & 20.46 & 17.77 \\
              & iCaRL & 94.85 & 72.71 & 63.24 & 54.69 & 54.1\\
              & UCIR & 94.89 & 80.16 & 71.22 & 57.26 & 59.91 \\
              & BiC & 93.37 & 72.09 & 60.87 & 57.89 & 55.23 \\
              & OneFi & 84.30 & 17.31 & 13.98 & 10.33 & 8.99 \\
              & Ours & 95.07 & 83.62 & 76.00 & 63.67 & 64.36 \\
        \midrule
        \multirow{5}{*}{mmWave} & Baseline & 91.11 & 35.2 & 27.49 & 20.47 & 17.32 \\
               & iCaRL & 91.70 & 73.22 & 64.6 & 55.7 & 55.82\\
               & UCIR & 91.41 & 76.36 & 72.0 & 62.21 & 61.79\\
               & BiC & 91.30 & 77.0 & 70.49 & 64.89 & 63.83\\
               & OneFi & 84.37 & 35.38 & 21.56 & 13.51 & 12.74 \\
               & Ours & 93.07 & 84.11 & 78.7 & 68.96 & 70.77\\
        \bottomrule
    \end{tabular}
    }
    \vspace{2pt}
    \caption{Accuracy of CCS services for user2 with the continuous demands listed in Table.~\ref{tab:user_sequences}. CCS outperforms existing alternative methods such as iCaRL~\cite{icarl}, UCIR~\cite{ucir}, BiC~\cite{bic} and OneFi~\cite{onefi} by a large margin for three wireless modalities. }
    \label{tab:result2}

\end{table}

\begin{figure}[t]
  \centering
  \includegraphics[width=\linewidth]{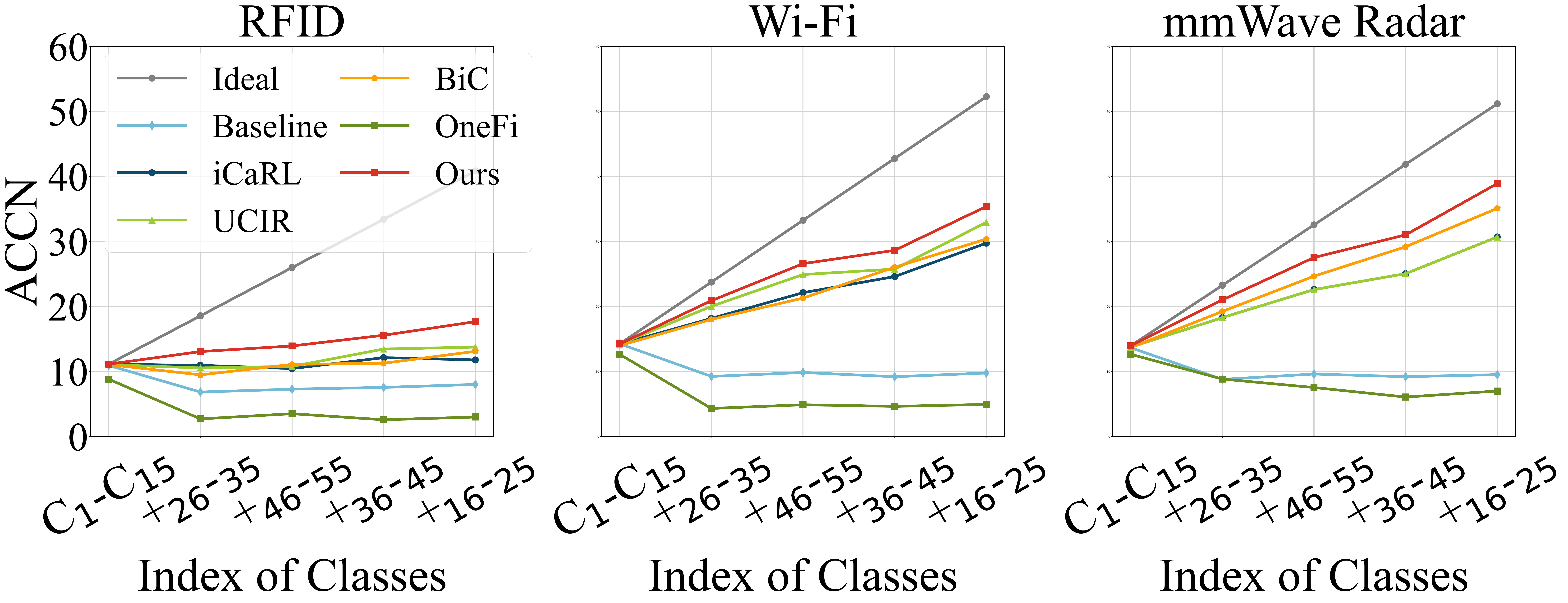}
  \caption{ ACCN of CCS services for user2 with the continuous demands listed in Table.~\ref{tab:user_sequences}. CCS traces the closest path to the ideal curve compared to alternative methods such as iCaRL~\cite{icarl}, UCIR~\cite{ucir}, BiC~\cite{bic} and OneFi~\cite{onefi}, demonstrating a consistent rise in model value. }
  \label{fig:result2}

\end{figure}

\begin{table}[t]
    \setlength{\tabcolsep}{2pt} 
    \centering
    \resizebox{\columnwidth}{!}{%
    \begin{tabular}{llccccc}
        \toprule
        \textbf{User3} & \textbf{Method} & \textbf{C1$\sim$C15} & \textbf{+C36$\sim$C45} & \textbf{+C46$\sim$C55} & \textbf{+C16$\sim$C25} & \textbf{+C26$\sim$C35} \\
        \midrule
        \multirow{5}{*}{RFID} & Baseline & 74.04 & 32.02 & 21.43 & 18.46 & 12.07\\
             & iCaRL & 73.74 & 41.47 & 30.03 & 28.37 & 23.28\\
             & UCIR & 73.70 & 49.53 & 34.76 & 31.19 & 24.64\\
             & BiC & 73.81 & 37.67 & 32.60 & 27.90 & 23.36\\
             & OneFi & 58.81 & 13.09 & 10.19 & 7.23 & 4.34 \\
             & Ours & 74.30 & 58.29 & 45.21 & 40.23 & 33.86\\
        \midrule
        \multirow{5}{*}{Wi-Fi} & Baseline & 95.07 & 37.38 & 28.3 & 21.85 & 17.14\\
              & iCaRL & 94.85 & 73.2 & 65.78 & 58.83 & 57.13\\
              & UCIR & 94.89 & 80.44 & 71.19 & 62.44 & 61.12\\
              & BiC & 93.37 & 73.71 & 62.95 & 60.12 & 54.03\\
              & OneFi & 84.30 & 18.78 & 13.95 & 11.04 & 6.79 \\
              & Ours & 95.07 & 83.4 & 78.38 & 66.72 & 63.62\\
        \midrule
        \multirow{5}{*}{mmWave} & Baseline & 91.11 & 36.8 & 27.43 & 21.35 & 16.16\\
               & iCaRL & 91.7 & 72.4 & 64.76 & 61.35 & 57.96\\
               & UCIR & 91.41 & 75.36 & 69.87 & 67.81 & 63.27\\
               & BiC & 91.3 & 75.02 & 70.22 & 69.27 & 65.24\\
               & OneFi & 84.37 & 28.24 & 21.76 & 15.56 & 9.05 \\
               & Ours & 93.07 & 84.64 & 78.41 & 71.70 & 69.78\\
        \bottomrule
    \end{tabular}
    }
    \vspace{2pt}
    \caption{Accuracy of CCS services for user3 with the continuous demands listed in Table.~\ref{tab:user_sequences}. CCS outperforms existing alternative methods such as iCaRL~\cite{icarl}, UCIR~\cite{ucir}, BiC~\cite{bic} and OneFi~\cite{onefi} by a large margin for three wireless modalities. }
    \label{tab:result3}
    
\end{table}

\begin{figure}[t]
  \centering
  \includegraphics[width=\linewidth]{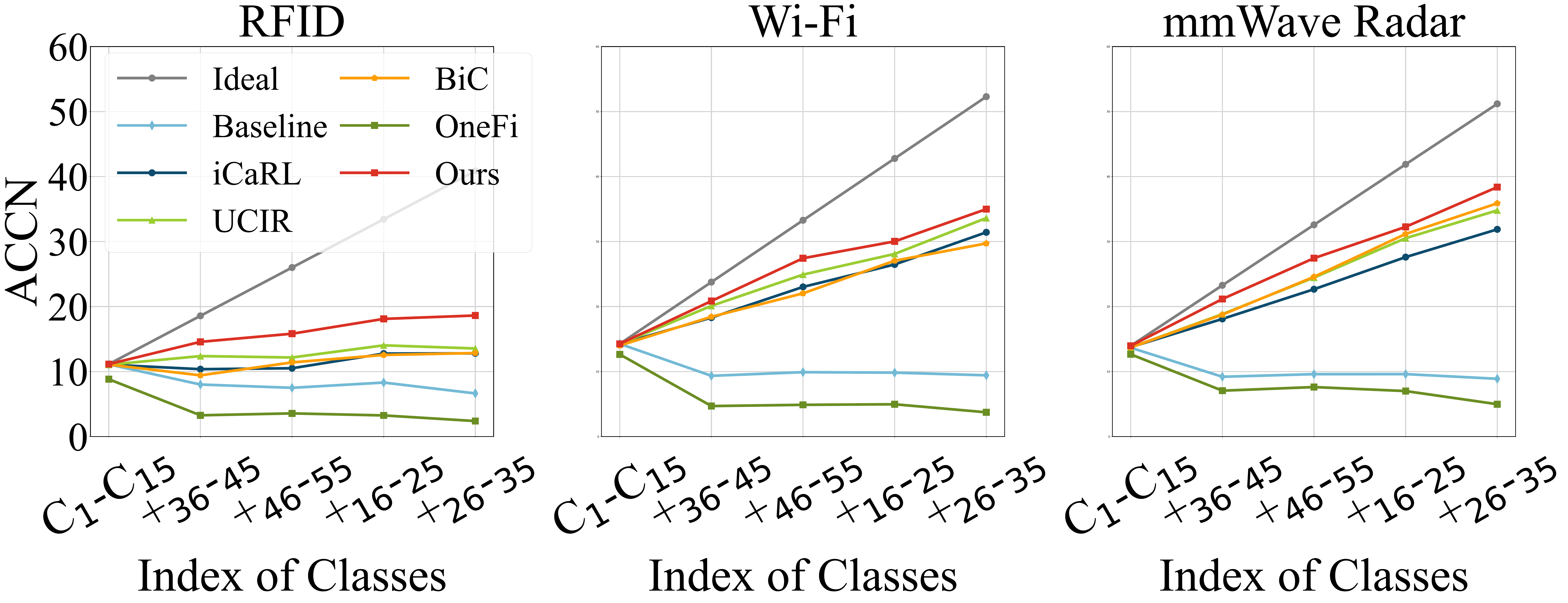}
  \caption{ ACCN of CCS services for user3 with the continuous demands listed in Table.~\ref{tab:user_sequences}. CCS traces the closest path to the ideal curve compared to alternative methods such as iCaRL~\cite{icarl}, UCIR~\cite{ucir}, BiC~\cite{bic} and OneFi~\cite{onefi}, demonstrating a consistent rise in model value. }
  \label{fig:result3}

\end{figure}

\textbf{(2) Accuracy and ACCN on User2}

In Table.~\ref{tab:result2}, we show CCS's accuracy on user2 at five continuous stages listed in Table.~\ref{tab:user_sequences}. Although CCS follows a different stage sequence for user2 compared to user1, which progresses incrementally in the order of C1$\sim$15, C26$\sim$C35, C46$\sim$C55, C36$\sim$C45, C16$\sim$C25, it consistently demonstrates high accuracy across all stages. By the end of the final stage, the accuracy rates for RFID, Wi-Fi, and millimeter-wave radar are 32.56\%, 63.87\%, and 67.45\%, correspondingly. Fig.~\ref{fig:result2} illustrates the ACCN curve of CCS services for user2. Our proposed method, CCS, traces the closest path to this ideal curve compared to current alternative methods like iCaRL~\cite{icarl}, UCIR~\cite{ucir}, and BiC~\cite{bic}, indicating a consistent rise in model value, measured by ACCN.

\textbf{(3) Accuracy and ACCN on User3}

In Table.~\ref{tab:result3}, we show CCS's accuracy on user3 at five continuous stages listed in Table.~\ref{tab:user_sequences}. Our method significantly mitigates models’ accuracy degradation. In final stage, our method achieves accuracy improvements of 9.22\%, 2.5\%, and 4.54\% in RFID, Wi-Fi and mmWave modalities respectively, compared to the second best method. Fig.~\ref{fig:result3} illustrates the ACCN curve of CCS services for user3. Our proposed solution can reap highest model value across all stages and modalites.

\textbf{(4) Accuracy and ACCN on User4}

\begin{figure}[t]
  \centering
  \includegraphics[width=\linewidth]{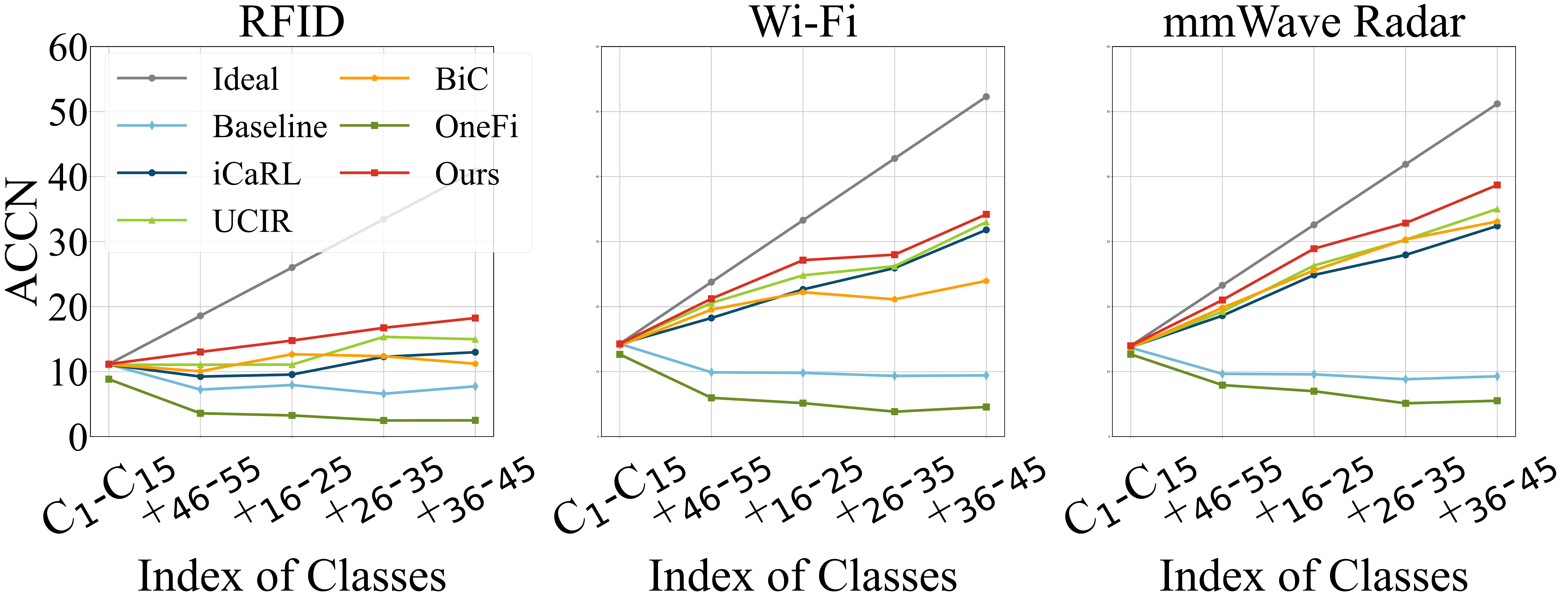}
  \caption{ ACCN of CCS services for user4 with the continuous demands listed in Table.~\ref{tab:user_sequences}. CCS traces the closest path to the ideal curve compared to alternative methods such as iCaRL~\cite{icarl}, UCIR~\cite{ucir}, BiC~\cite{bic} and OneFi~\cite{onefi}, demonstrating a consistent rise in model value. }
  \label{fig:result4}
\end{figure}

In Table.~\ref{tab:result4}, we show CCS's accuracy on user4 at five continuous stages listed in Table.~\ref{tab:user_sequences}. Our method significantly mitigates models’ accuracy degradation. In the final stage, our method achieves accuracy improvements of  5.92\%,2.16\% and 6.71\% in RFID, Wi-Fi, and mmWave modalities, respectively, compared to the second best method. Fig.~\ref{fig:result4} illustrates the ACCN curve of CCS services for user4. Our proposed solution has the potential to yield the highest model value across all stages and modalities.

The results in Sec.~\ref{sec:results} show that CCS consistently performs best across different incremental demand sequences for four users. Additionally, it demonstrates increasing model value over time, highlighting CCS's versatility in addressing customized incremental wireless sensing services.

\begin{table}[t]
     \setlength{\tabcolsep}{2pt} 
    \centering
    \resizebox{\columnwidth}{!}{%
    \begin{tabular}{llccccc}
        \toprule
        \textbf{User4} & \textbf{Method} & \textbf{C1$\sim$C15} & \textbf{+C46$\sim$C55} & \textbf{+C16$\sim$C25} & \textbf{+C26$\sim$C35} & \textbf{+C36$\sim$C45} \\
        \midrule
        \multirow{5}{*}{RFID} & Baseline & 74.04 & 28.91 & 22.65 & 14.63 & 14.07\\
             & iCaRL & 73.74 & 36.93 & 27.25 & 27.27 & 23.59\\
             & UCIR & 73.70 & 44.20 & 31.63 & 34.09 & 27.23\\
             & BiC & 73.81 & 40.18 & 36.19 & 27.47 & 20.36\\
             & OneFi & 58.81 & 14.31 & 9.30 & 5.51 & 4.53 \\
             & Ours & 74.30 & 52.07 & 42.21 & 37.17 & 33.15\\
        \midrule
        \multirow{5}{*}{Wi-Fi} & Baseline & 95.07 & 39.53 & 27.98 & 20.74 & 17.12\\
              & iCaRL & 94.85 & 72.96 & 64.68 & 57.62 & 57.81\\
              & UCIR & 94.89 & 82.18 & 70.9 & 58.23 & 60.01\\
              & BiC & 93.37 & 78.11 & 63.46 & 46.91 & 43.55\\
              & OneFi & 84.30 & 23.84 & 14.70 & 8.51 & 8.28 \\
              & Ours & 95.07 & 84.82 & 77.54 & 62.19 & 62.17\\
        \midrule
        \multirow{5}{*}{mmWave} & Baseline & 91.11 & 38.60 & 27.35 & 19.60 & 16.86\\
               & iCaRL & 91.70 & 74.38 & 71.02 & 62.10 & 58.91\\
               & UCIR & 91.41 & 76.96 & 75.17 & 67.31 & 63.66\\
               & BiC & 91.30 & 79.22 & 73.06 & 67.38 & 60.16\\
               & OneFi & 84.37 & 31.69 & 19.94 & 11.37 & 10.02 \\
               & Ours & 93.07 & 84.00 & 82.59 & 72.98 & 70.37\\
        \bottomrule
    \end{tabular}
    }
    \vspace{2pt}
    \caption{Accuracy of CCS services for user4 with the continuous demands listed in Table.~\ref{tab:user_sequences}. CCS outperforms existing alternative methods such as iCaRL~\cite{icarl}, UCIR~\cite{ucir}, BiC~\cite{bic} and OneFi~\cite{onefi} by a large margin for three wireless modalities. }
    \label{tab:result4}
      
\end{table}

\subsection{Ablation Study}\label{sec:ablation}
In this section, we first conduct ablation study on the effectiveness of the combination among the applied techniques in CCS, including exemplar selection, knowledge distillation and weight aligning. Besides, we evaluate different loss functions in knowledge distillation and different normalization method in weight aligning. 

\begin{table}[t]
    \small
    \centering
     \setlength{\tabcolsep}{2pt} 
    \resizebox{\columnwidth}{!}{%
    \begin{tabular}{lcccccccc}
    \toprule
    \multirow{2}{*}{\textbf{Modality}} & \multicolumn{3}{c}{\textbf{Method}}       & \multicolumn{1}{c}{\multirow{2}{*}{C1$\sim$C15}} & \multicolumn{1}{c}{\multirow{2}{*}{+C16$\sim$C25}} & \multicolumn{1}{c}{\multirow{2}{*}{+C26$\sim$C35}} & \multicolumn{1}{c}{\multirow{2}{*}{+C36$\sim$C45}} & \multicolumn{1}{c}{\multirow{2}{*}{+C46$\sim$C55}} \\
    
    & \multicolumn{1}{c}{\textbf{E}} & \multicolumn{1}{c}{\textbf{KD}} & \multicolumn{1}{c}{\textbf{WA}} & \multicolumn{1}{c}{} & \multicolumn{1}{c}{} & \multicolumn{1}{c}{} & \multicolumn{1}{c}{} & \multicolumn{1}{c}{}\\
    \midrule
    \multirow{5}{*}{RFID} 
        & / & \checkmark & \checkmark & 74.04 & 33.24 & 23.06 & 19.41 & 14.72\\
        & \checkmark & / & / & 74.04 & 43.53 & 28.33 & 28.07 & 20.13\\
        & \checkmark & \checkmark & / & 74.30 & 48.20 & 33.75 & 28.59 & 22.87\\
        & \checkmark & / & \checkmark & 74.30 & 48.80 & 35.41 & 37.27 & 30.06\\
        & \checkmark & \checkmark & \checkmark & 75.96 & 58.56 & 44.89 & 38.23 & 34.01\\
    \midrule
    \multirow{5}{*}{Wi-Fi} 
        & / & \checkmark & \checkmark & 95.07 & 41.53 & 33.76 & 25.98 & 24.20\\
        & \checkmark & / & / & 95.07 & 72.00 & 65.57 & 52.22 & 56.01\\
        & \checkmark & \checkmark & / & 94.85 & 73.38 & 66.89 & 55.98 & 56.87\\
        & \checkmark & / & \checkmark & 95.07 & 76.18 & 69.40 & 56.95 & 61.08\\
        & \checkmark & \checkmark & \checkmark & 95.07 & 83.56 & 75.48 & 64.12 & 65.10\\
    \midrule
    \multirow{5}{*}{mmWave}
        & / & \checkmark & \checkmark & 91.11 & 62.38 & 47.86 & 41.20 & 35.29\\
        & \checkmark & / & / & 93.07 & 76.38 & 66.21 & 57.96 & 56.14\\
        & \checkmark & \checkmark & / & 93.07 & 84.07 & 73.97 & 62.74 & 59.45\\
        & \checkmark & / & \checkmark & 93.07 & 80.91 & 73.24 & 60.72 & 58.25\\
        & \checkmark & \checkmark & \checkmark & 93.07 & 87.49 & 81.16 & 69.35 & 69.19\\
    \bottomrule
    \end{tabular}
    }
    \vspace{2pt}
    \caption{Accuracy of CCS services in ablation study for user1 with the continuous stages listed in Table.\ref{tab:user_sequences}. E, KD, WA represent Herding exemplar selection, knowledge distillation and  weight aligning respectively.}
    \label{tab:ablation_ekdwa}
    
\end{table}

\begin{figure}[t]
  \centering
  \includegraphics[width=\linewidth]{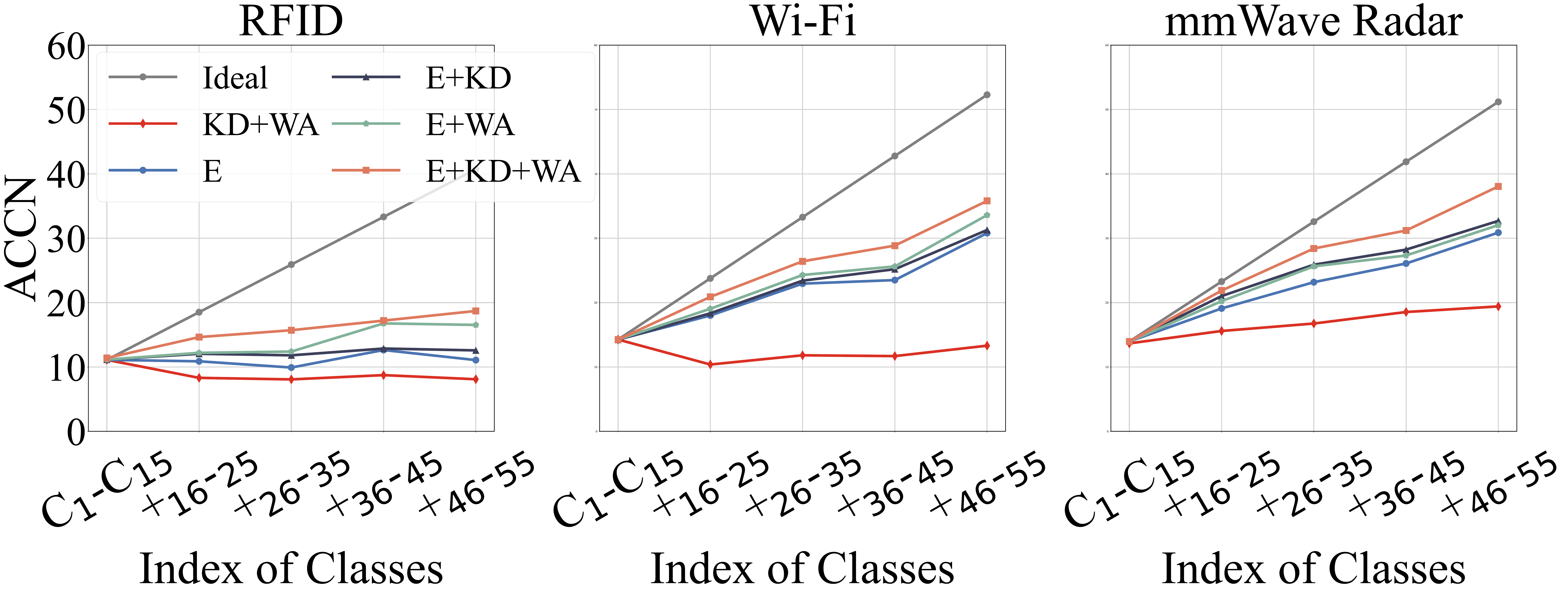}
  \caption{ACCN of CCS services in ablation study for user1 with the continuous stages listed in Table.\ref{tab:user_sequences}. }
  \label{fig:ablation_ekdwa}
\end{figure}

\textbf{(1) Combination of Exemplar, Knowledge Distillation, and Weight Aligning}

We conduct the ablation experiment based on user1 at five continuous stages listed in Table.~\ref{tab:user_sequences}. Table.~\ref{tab:ablation_ekdwa} shows that exemplar has a significant impact on reducing model's performance degradation. Without the assistance of exemplar, the accuracy will be 19.29\%,40.9\% and 33.9\% lower at final stage in RFID, Wi-Fi and mmWave modalities respectively, even if we adopt knowledge distillation and weight aligning techniques. Based on exemplar, we apply knowledge distillation and weight aligning techniques separately. As shown in Table.~\ref{tab:ablation_ekdwa}, weight aligning can increase accuracy of model to a greater extend in RFID and Wi-Fi modalities, and to a lower extend in mmWave modality than knowledge distillation with the assistance of exemplar. Moreover, using knowledge distillation, weight aligning and exemplar at the same time can reap better results than using them separately. Fig.~\ref{fig:ablation_ekdwa} illustrates the ACCN curve for ablation study. When we train model using knowledge distillation and weight aligning without exemplar, the ACCN values of models maintain at low values at all stages. In RFID and Wi-Fi modalities, the ACCN values are even lower than values at initial stage.  Our method, which utilizes knowledge distillation,weight aligning and exemplar at the same time, keeps on increasing model value as we introduce more unseen classes. Eliminating any techniques in our method can result in lower ACCN values.

\textbf{(2) Loss that applied in Knowledge Distillation}

\begin{figure}[t]
  \centering
  \includegraphics[width=\linewidth]{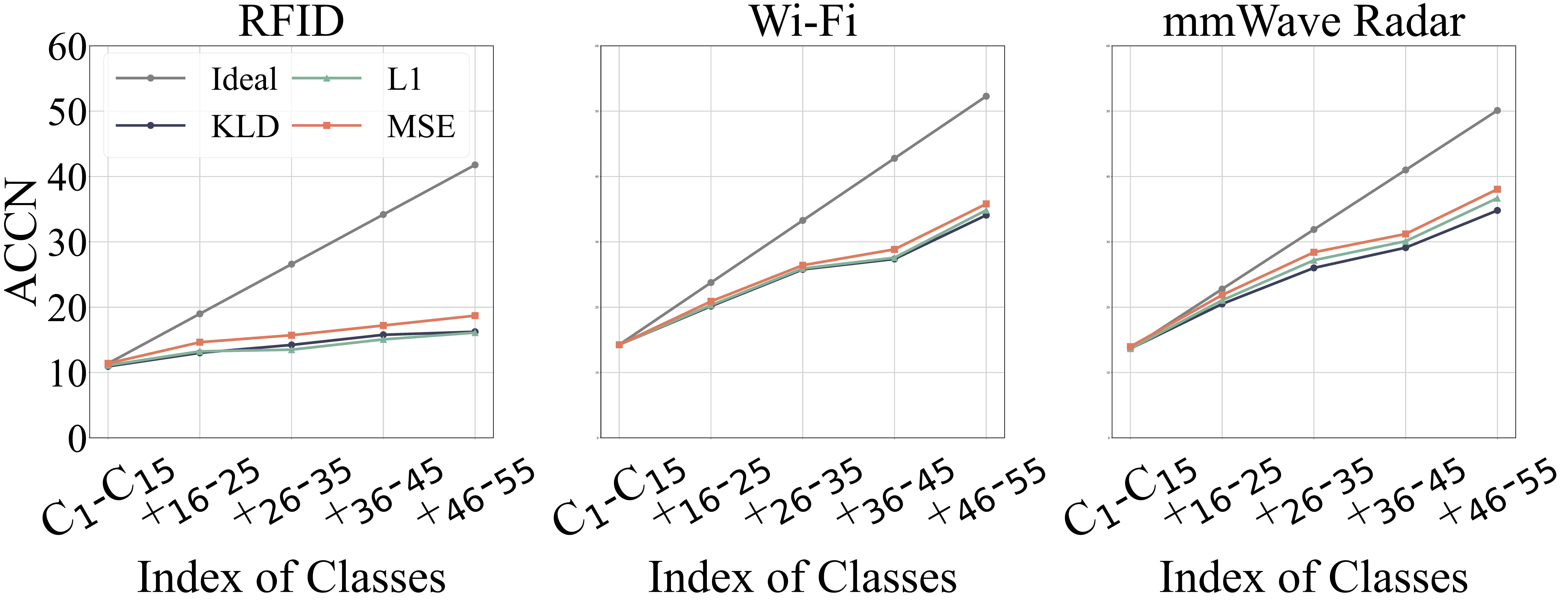}
  \caption{Comparative analysis of CCS services' ACCN for user1 with the continuous stages listed in Table.\ref{tab:user_sequences} on different distillation losses, including Kullback Leibler~(KL) Divergence distance loss, L1 loss and mean square error loss.}
  \label{fig:ab_loss}
\end{figure}

We compare performance of knowledge distillation based on different loss functions.Table.~\ref{tab:ab_loss} lists accuracy for user1 at five continuous stages listed in Table.~\ref{tab:user_sequences} when implementing knowledge distillation using Kullback Leibler~(KL) Divergence distance loss, L1 loss, and mean square error loss (MSE). 
The results shown in Table.~\ref{tab:ab_loss} indicate that knowledge distillation based on mean square error loss provides best results among these three losses. As Fig.~\ref{fig:ab_loss} illustrated, all three losses can effectively increase model value. Mean-square error loss can receive slightly higher model value than the other two losses in RFID, Wi-Fi and mmWave modalites. Therefore, we select mean-square error loss as the distillation loss of our method.
\begin{table}[t]
     \setlength{\tabcolsep}{2pt} 
    \centering
    \resizebox{\columnwidth}{!}{%
    \begin{tabular}{llccccc}
        \toprule
        \textbf{Modality} & \textbf{Method} & \textbf{C1$\sim$C15} & \textbf{+C16$\sim$C25} & \textbf{+C26$\sim$C35} & \textbf{+C36$\sim$C45} & \textbf{+C46$\sim$C55} \\
        \midrule
        \multirow{3}{*}{RFID} & KLD & 72.96 & 52.07 & 40.65 & 35.06 & 29.53\\       & L1 & 74.04 & 52.98 & 38.56 & 33.49 & 29.29\\
             & MSE & 75.96 & 58.56 & 44.89 & 38.23 & 34.01\\
        \midrule
         \multirow{3}{*}{Wi-Fi}  & KLD & 95.07 & 80.64 & 73.67 & 60.84 & 61.97\\
                & L1 & 95.07 & 81.24 & 74.06 & 61.28 & 63.37\\
                & MSE & 95.07 & 83.56 & 75.48 & 64.12 & 65.10 \\
        \midrule
        \multirow{3}{*}{mmWave} & KLD & 91.11 & 82.0 & 74.32 & 64.69 & 63.29\\
               & L1 & 91.11 & 84.24 & 77.67 & 66.89 & 66.70\\
               & MSE & 91.11 & 84.31 & 77.84 & 69.19 & 68.54\\
        \bottomrule
    \end{tabular}
    }
    \vspace{2pt}
    \caption{Comparative analysis of CCS services' accuracy for user1 with the continuous stages listed in Table.\ref{tab:user_sequences} on different distillation losses, including Kullback Leibler~(KL) Divergence distance loss, L1 loss and mean square error loss.}
    \label{tab:ab_loss}

\end{table}

\textbf{(3) Normalization Method in Weight Aligning}

We compare accuracy and ACCN of CCS services for user1 at five continuous stages listed in Table.~\ref{tab:user_sequences} when using L1 norm and L2 norm to implement weight aligning technique. As shown in Table~\ref{tab:ab_norm}, L2 norm outperforms L1 norm at all stages. The accuracy will be 7.84\%,0.39\% and 1.39\% higher at final stage in RFID, Wi-Fi and mmWave modalities respectively when using L2 norm instead of L1 norm. 

\begin{table}[t]
     \setlength{\tabcolsep}{2pt} 
    \centering
    \resizebox{\columnwidth}{!}{%
    \begin{tabular}{llccccc}
        \toprule
        \textbf{Modality} & \textbf{Method} & \textbf{C1$\sim$C15} & \textbf{+C16$\sim$C25} & \textbf{+C26$\sim$C35} & \textbf{+C36$\sim$C45} & \textbf{+C46$\sim$C55} \\
        \midrule
        \multirow{2}{*}{RFID} & L1 & 74.04 & 52.31 & 38.37 & 29.98 & 26.17 \\
             & L2 & 75.96 & 58.56 & 44.89 & 38.23 & 34.01\\
        \midrule
        \multirow{2}{*}{Wi-Fi} & L1 & 95.07 & 82.31 & 75.11 & 62.73 & 64.71\\
               & L2 & 95.07 & 83.56 & 75.48 & 64.12 & 65.1\\
        \midrule
        \multirow{2}{*}{mmWave} & L1 & 91.11 & 84.22 & 77.67 & 68.11 & 67.15\\
               & L2 & 91.11 & 84.31 & 77.84 & 69.19 & 68.54\\
        \bottomrule
    \end{tabular}
    }
    \vspace{2pt}
    \caption{Comparative analysis of CCS services' accuracy for user1 with the continuous stages listed in Table.\ref{tab:user_sequences} when using L1 norm and L2 norm to implement weight aligning technique.}
    \label{tab:ab_norm}

\end{table}



Fig.\ref{fig:ab_loss} presents ACCN values of CCS services when using different norms for weight aligning. The ACCN curves in Wi-Fi and mmWave modalities exhibit remarkable similarity, while ACCN curve for L2 norm approaches the ideal curve more closely than ACCN curve for L1 norm in RFID modality. The results indicate that weight aligning technique based on L2 norm can assist in maintaining models' performance better than technique based on L1 norm. Therefore, our method apply L2 norm to implement weight aligning technique.

\section{Conclusion}\label{sec:conclusion}
Perceiving wireless sensing as being at the tipping point of transitioning from proof-of-concept to large-scale deployment, we take a step forward by envisioning a service schema. This model entails wireless sensing service providers and users, where providers offer sensing models to users who, in turn, may continuously generate new sensing demands through usage. To enable providers to meet these ongoing user needs, we propose CCS, which facilitates the provision of updated models to satisfy emerging demands. Incorporating techniques like continuous learning, exemplar selection, knowledge distillation, and weight aligning, CCS ensures that new models not only meet new demands but also retain capability for old services. Extensive experiments on the large-scale XRF55 dataset, including sequential introduction of new demands, comparative studies, and ablation experiments, demonstrate the superior sustained service capability of CCS, significantly outperforming existing approaches. Furthermore, we delve into CCS by discussing it from multiple views, including user acceptance, provider incentives, model integration, model capacity, and federated learning. 

While CCS may not be the sole or perfect model for future wireless sensing services, it serves as a catalyst for academia and industry to contemplate the paradigms of service or cloud services beyond this tipping point.

\bibliographystyle{IEEEtran}
\bibliography{reference.bib}


\end{document}